# Disrupting Adversarial Transferability in Deep Neural Networks

Christopher Wiedeman[1], Ge Wang*[2]

## Summary

Adversarial attack transferability is well-recognized in deep learning. Prior work has partially explained transferability by recognizing common adversarial subspaces and correlations between decision boundaries, but little is known beyond this. We propose that transferability between seemingly different models is due to a high linear correlation between the feature sets that different networks extract. In other words, two models trained on the same task that are distant in the parameter space likely extract features in the same fashion, just with trivial affine transformations between the latent spaces. Furthermore, we show how applying a feature correlation loss, which decorrelates the extracted features in a latent space, can reduce the transferability of adversarial attacks between models, suggesting that the models complete tasks in semantically different ways. Finally, we propose a Dual Neck Autoencoder (DNA), which leverages this feature correlation loss to create two meaningfully different encodings of input information with reduced transferability.

**Keywords**: Deep Learning, Computer Vision, Adversarial Attacks, Attack Transferability, Decorrelation

## Introduction

The advancement of artificial intelligence, especially deep learning (DL), has revolutionized research in computer vision, natural language processing, and recently many other fields, including medical image reconstruction. Unfortunately, the development of adversarial attacks against deep learning models has highlighted a glaring lack of robustness in DL methods. Adversarial attacks to deep neural networks were first discussed in 2014 when Szegedy, et al. showed that adding deliberately crafted but imperceivable perturbations to MNIST digit images would cause a trained classifier to catastrophically misclassify the sample.[1] Since then, successful adversarial attacks have been demonstrated in numerous other DL contexts, such as speech-to-text translation and medical image reconstruction.[2,3] Counterintuitively, networks that can generalize to never-seen natural samples cannot perform well on samples very close to previously seen samples. Such a discovery not only raises security concerns but also questions whether DL models truly learn the desired features for a given task.

Initially, it was thought that adversarial samples were a product of overfitting, and partially due to the capacity for deep neural networks to fit overly-complex decision boundaries, but lower capacity networks and less complex machine learning models are more vulnerable to adversarial attacks.[4–6] On the contrary, the current thinking is that data-driven models (particularly more linear models) only learn samples along a data manifold within the input space, whereas adversarial examples exist off-manifold. Complex, highly non-linear networks with sufficient training data are required to expand this manifold, which becomes increasingly challenging in high-dimensional input spaces.

[1]Rensselaer Polytechnic Institute, Department of Electrical and Computer Systems Engineering, Troy, NY, USA
[2]Rensselaer Polytechnic Institute, Department of Biomedical Engineering, Troy, NY, USA
*Correspondence: wangg6@rpi.edu

For a comprehensive review on the research in adversarial attacks, we recommend reading *Adversarial Attacks and Defenses in Deep Learning* by Ren, et al.[7]

*Types of attacks:* An untargeted adversarial attack is one that does not seek a specific result (e.g., misclassification into a specific class) aside from maximally degrading a performance measure of the model. If $f$ is a trained model, with a loss objective $J(\theta, x, y)$ where $\theta$ are the model parameters and $\{x, y\}$ are data inputs and labels respectively, then an untargeted attack based on sample/label pair $\{x_i, y_i\}$ finds point $x_i'$ near $x_i$ that maximizes the loss:

$$x_i' = \underset{x_i'}{\operatorname{argmax}} J(\theta, x_i', y_i), \quad D(x_i, x_i') \leq \epsilon \tag{1}$$

where $D$ is some distance metric. In the following work, we use the norm $L_\infty(x_i - x_i')$ (unless otherwise specified) for $D$, as this is the most used distance metric in adversarial attacks, although $L_0$, $L_1$, and $L_2$ are also occasionally used. Numerous methods have been discovered to find adversarial attacks, many of which rely on first-order optimization. In this work, we will primarily focus on two common search methods: Fast Gradient Sign Method (FGSM)[5] and Projected Gradient Descent (PGD).[6,8] FGSM is favored for its simplicity and efficiency:

$$x' = x + \epsilon \operatorname{sgn}(\nabla_x J(\theta, x, y)) \tag{2}$$

On the other hand, PGD is like FGSM, but uses a random initialization and iteration to find stronger attacks. Clipping is used to keep the sample within the distance constraint and data range:

$$x^{t+1} = \operatorname{Clip}(x^t + \alpha \operatorname{sgn}(\nabla_x J(\theta, x, y))) \tag{3}$$

where $\alpha$ is some step size. PGD is considered a strong attack, and attaining robustness against this method will generally defend against other attacks.[6]

Adversarial attacks are referred to as either white, gray, or black box attacks to specify the attacker's knowledge on the target model. In white box attacks, the attacker has full access to the model's architecture and parameters. In gray box attacks, the attacker only knows the model architecture, and black box attackers assume no information on the target model. In both gray and black box scenarios, an attacker typically trains a surrogate model as the target model, and then crafts attacks using this surrogate. While white box attacks are understandably the most potent, gray and black box attacks have been also shown potency due to high attack transferability between models.

*Related Work*: To our best knowledge, Szegedy et al. raised the first concern regarding adversarial stability of DNNs in 2014.[1] A plethora of work proposing both new attack and defense methods has been published since then. Carlini and Wagner have contributed much research to stronger attack methods, including the creation of C&W attacks that are effective against many earlier defenses such as defensive distillation.[9–11] Papernot, et al. also proposed a class of algorithms that use an adversarial saliency map to produce targeted (i.e., producing a specific result) attacks.[12] A few works also used generative adversarial networks to produce attacks.[13,14]

On the defense side, it is crucial to distinguish between adversarial defenses and true adversarial robustness. Many adversarial defenses achieve success against specific, often weaker attacks, but seemingly fail against stronger attacks. It was discovered that many of these methods defend against unsophisticated attacks because they obfuscate the loss gradient, making effective attacks harder to find.[15] In this sense, the network is secure against weak attacks but not truly robust, since adversarial examples still exist, and can be exploited with stronger algorithms. Currently, the only approach that has demonstrated true robustness is adversarial training, in which a network is trained with adversarial examples. Numerous methods, including GAN-based adversarial sample generation have been explored.[5,6,14,16] Robustness against strong attacks, however, requires training on strong examples, which are prohibitively expensive in most contexts, and high-capacity networks for fitting complex decision boundaries.[6]

An intriguing property of adversarial attacks is their transferability between models, i.e., adversarial samples trained to target one model are often successful against other models, even when these models are trained with different data or use different architectures.[1] It has been shown that DNNs can be robust to random perturbations (e.g., Gaussian noise) while still being highly vulnerable to adversarial attacks.[1] This suggests that while many adversarial examples may exist, the total adversarial space is not random and still small relative to the high dimensional input space; the intersection of these spaces between models cannot be attributed to mere coincidence. As such, numerous works have rationalized the transferability phenomenon by pointing out correlated decision boundaries for samples between different models.[5,17–19] There is indeed strong evidence that the decision boundaries around samples are correlated between different models, and that adversarial samples exist in subspaces that often intersect.[18] This phenomenon is clearly observed in *universal adversarial perturbations*, which generalize not only between models, but also between different input samples due to strong correlations between decision boundaries.[19,20] Still, this explanation does not explain *why* these decision boundaries and subspaces are correlated between models with entirely different parameters or architectures.

How does one train meaningfully different models and avoid attack transferability? A naïve approach to this question would maximize the distance between two models in the parameter space, as it is known that DNNs typically do not converge to a global minimum, but rather approach one of many local minima. This method understandably assumes that differently parameterized models solve a task in different ways. In this paper, we unpeel one more layer of the transferability puzzle by first demonstrating that DNNs trained on the same task are quite similar in the way they complete tasks, regardless of how different the models are in the parameter space. In fact, we find that the features that two models extract are semantically identical, and with one simply being an affine-transformed version of the other. More importantly, we show that breaking this correlation drastically reduces the transferability of adversarial samples between models. Finally, we illustrate this concept of breaking transferability with a dual-neck autoencoder (DNA), which creates two unique input encodings (also referred to as bottlenecks) with a limited number of layers added to a traditional autoencoder.

# Results

## Transferability and Feature Correlation between Models

First, we observed how parametric distance between models affects attack transferability. Convolutional neural networks (CNNs) of identical architecture were trained in parallel for MNIST classification with varying distances in the parameter space. Figure 1 reports adversarial attack transferability rates between these model pairs. We found that transferability unsurprisingly increased with attack magnitude, but there was no correlation at all between transferability and distance in the parameter space, contradicting the notion that parametrically distant networks solve tasks in different ways.

To investigate this further, we compared one of the late feature representations between model pairs. We found that for various sample batches $X$, the Pearson correlation coefficient between the two hidden states produced by the models was $< 0.975$ on average. This high degree of correlation not only held for natural samples, but also for adversarial samples, and was true regardless of the parameter distance between the two models (Figure 1), revealing a strong, consistent linear relationship between the two encodings. This implies that the information extracted by the two models is quite similar, with encodings only separated by an affine transform.

## Disrupting Transferability through Decorrelating Features

### Key idea

Given such a linear correlation between features extracted by networks, a natural question is how networks would behave if this correlation were disrupted. We test this scenario by once again training two classifying CNNs in parallel but introduce a decorrelating term in the training objective, which decouples latent features extracted by the two models. This decorrelating term, which we refer to as correlation loss, is related to the correlation coefficient of features for a sample batch at a proper hidden layer in the neural networks (implementation details in Experimental Procedures). Figure 2 illustrates the effect of this correlation loss on the compressed sample features at this hidden layer: while the compressed features extracted by two parallel models are normally well correlated, it can be seen that introducing correlation loss significantly reduces this correlation. Note that the feature space for each model was compressed to a single dimension using principal component analysis (PCA).

To evaluate the effects of feature decorrelation on attack transferability, transfer rates between model pairs were evaluated with varying factors, including attack algorithm ($FGSM\ \ell_\infty, FGSM\ \ell_2, PGD\ \ell_\infty, PGD\ \ell_2$), model architecture (CNN based of fully-connected with batch normalization), and attack strength ($\epsilon = 0.05, 0.10, 0.15, 0.2$). For $\ell_2$ attacks, we modified the traditional FGSM and PGD algorithms to appropriately be constrained by $\ell_2$ norms. Figure 3 compares the transferability of the trained model pairs with and without use of the correlation loss under these conditions. We found that applying the correlation loss led to a considerable reduction in adversarial transferability, although, this difference diminished at higher magnitudes, perhaps because the noise level effectively generated out-of-distribution samples. Furthermore, we found that imposing this decorrelation constraint minimally impacted the final accuracy on natural samples.

### Combining Decorrelation with DVERGE

More recently, a newer perspective on adversarial instability has arisen, viewing it as a byproduct of supervised learning that prioritizes well-generalizing features. Specifically, Ilyas, et al. hypothesize that in the current DNN paradigm, models identify subtle, non-robust features that paradoxically generalize well over the entire data distribution, and these features are easily disrupted with small, targeted adversarial noise.[21] Transferability among models exists because the same non-robust features are learned during the training process. Yang, et al. attempt to train an ensemble of diverse models to learn different non-robust features, distributing vulnerability and reducing attack transferability. They refer to this method as DVERGE.[22] We suspect that our method of feature decorrelation also reduces transferability by diversifying the non-robust features learned. While DVERGE takes a more direct approach by searching for distilled class features, the approach, much like adversarial training is computationally expensive as it involves iterative optimization of new samples. As such, it is perhaps logical to use a decorrelation mechanism to create initial diversity between models, and then 'fine-tune' this diversity using a DVERGE approach. To test this, we apply the principle of DVERGE training to model pairs initially trained with and without correlation loss. The results of this experiment (Figure *4*), show that using decorrelation in initial model training benefits later DVERGE training, as the models are already trained to be more diverse, resulting in lower attack transfer rates.

## Dual-Neck Autoencoder (DNA)

### Key idea

To further explore the effect of the correlation loss in training parallel models, we propose a Dual Neck Autoencoder (DNA), which consists of a traditional autoencoder/decoder structure that diverges into two paths at the most compressed hidden layer (i.e., the bottleneck). The most basic motivating concept behind this idea, commonly found in nature and communication systems, is information redundancy: two differently processed but overlapping information representations are encoded and decoded in parallel. The architecture contains shared encoding and decoding pathways (otherwise, the aggregate will essentially be two entirely separate models), yet the pathways must be meaningfully different, lest both pathways will only contain trivial differences (spatial rotations, shifts, etc.) and will suffer from identical instabilities.

Figure 5 illustrates the general structure of our proposed DNA. A shared encoder block first processes and partially compresses the input. This encoding is then fed into two bottleneck modules, each of which further compresses the information before partially decompressing. During training, the latent representations at the most compressed state are compared between the two bottlenecks using the correlation loss. The two partial decompressions are then both fully decompressed with a common decoder, producing two separate reconstructions. In the experiments, a trained classifier is also stacked at the end of this module to evaluate the reconstructions.

Both the MNIST Digit and CIFAR-10 datasets were used to test DNA. Simple fully connected architectures for MNIST and convolutional architectures for CIFAR-10 were designed for the encoder, bottleneck, and decoder blocks shown in Figure 5 (exact architectures in Experimental

Procedures). In both cases, the DNA was first trained for reconstruction fidelity with the following loss function:

$$J_{DNA}(F, Z, X_i) = \text{MSE}(F(X_i, \boldsymbol{\theta})_1, X_i)^2 + \text{MSE}(F(X_i, \boldsymbol{\theta})_2, x_i)^2 + \lambda \mathcal{L}_R(F, \boldsymbol{\theta}, X_i) \qquad (4)$$

where MSE is the mean-square error between two arguments, $F(\cdot)_n$ is the $n_{th}$ reconstruction pathway (e.g., $encoder \rightarrow bottleneck_n \rightarrow decoder$), $X_i$ is an input batch of images, and $\boldsymbol{\theta} = [\theta_{encoder}, \theta_{bottleneck1}, \theta_{bottleneck2}, \theta_{decoder}]$ are the network parameters. Correlation loss $\mathcal{L}_R(F, \boldsymbol{\theta}, X_i)$ uses the compressed bottleneck representations for correlation comparison. Note that if one disregards the correlation loss, this is the squared $L_2$ norm of the mean-square errors of the two reconstructions, which is preferred to avoid a 'sparse' solution (i.e., avoid the optimizer only improving one reconstruction at the expense of the other). A classifier was then attached and trained to the end of the DNA, and classification-based adversarial attacks were designed to target both reconstructions simultaneously.

Figure 6 illustrates classification metrics for DNA reconstructions of MNIST digit data under FGSM and PGD attacks. A sample reconstruction was considered accurate if one or both output reconstructions were correctly classified. Additionally, the average accuracy for each DNA pathway considered individually is recorded. For comparison, a traditional autoencoder with identical encoder, bottleneck, and decoder architectures was trained and evaluated in a similar fashion. Unsurprisingly, all reconstruction accuracies decreased with higher magnitude attacks, with PGD attack being more potent. As expected, the DNA structure was able to protect more samples compared to the traditional autoencoder.

Figure 7 shows five examples of digit reconstructions from adversarial examples. Qualitatively, one can see that the attack causes a semantic corruption in one of the bottleneck encodings, but this corruption does not transfer to the alternative bottleneck, despite sharing the same encoder/decoder layers.

Figure 8 illustrates accuracy metrics for FGSM and PGD attacks for the CIFAR-10 reconstructions. Classification accuracies are overall lower than the results seen in MNIST experiments, but the overall trends are similar, with a significant performance gap between the DNA and traditional autoencoder. Figure *9* also shows sample reconstructions of adversarial examples, but the effects of corruption on the reconstruction process are less clear.

To investigate the hyperparameters that influence transferability in the DNA, we made various architectural changes to the baseline model and obtained the results. While we found that increasing the width and depth of the baseline bottleneck did not always significantly decrease transferability by itself, we instead found that altering the proportion of shared/unshared layers influenced transferability considerably. In the baseline model, each reconstruction pathway consists of six sequential layers, in which all but the middle two layers are shared (Figure 12). We retested the model, expanding this unshared portion to four layers and all six layers (i.e., two separate autoencoder models in the latter case), and observed the resulting accuracy in Figure 10. Clearly, increasing the proportion of unshared layers consistently increased the likelihood of producing an accurate reconstruction.

## Discussion

The results displayed in Figure 1 suggest a striking revelation: the latent features extracted by two models with vastly disparate parameters are numerically different but encode the same semantic information. Furthermore, the relationship between the two encodings is linear, with one being an affine transformation of the other. This also suggests that DNNs distant in the parametric space are not necessarily semantically different; rather, they can (and often do) encode the same information, only differing by trivial transformations in their latent spaces (e.g., rotations, expansions/contractions, shifts). This insight is in line with observations from,[1,23] which note that individual units in latent spaces do not correspond to useful information; rather, vectors in the space as a whole represent features. It also helps explain why optimizing for a global minimum is not necessary, and why numerous local minima with similar loss values exist in the parameter space, as these points do not represent semantically different models. It is no wonder that these 'different' neural networks perform similarly on adversarial examples. To enforce the relationship between this feature correlation and adversarial attack transferability, Figure 1 also illustrates that this correlation holds with adversarial samples, indicating that the different networks respond to adversarial attacks in a predictably similar fashion, and Figure 2 reveals that breaking this correlation drastically reduces attack transferability. It is possible that this reduction in attack transferability is due to a diversification in the non-robust features learned by each model.[21] As such, decorrelation could be used as a tool to create diversify learned features between models, and Figure 4 suggests it as a useful pretraining tool for diverse ensembles such as DVERGE.[22]

Overall, results in both the MNIST and CIFAR-10 experiments indicate a reconstruction stability benefit from the decorrelated dual-bottlenecks in the proposed DNA network. This is seen at all tested points in Figure 6 and Figure 8, where the DNA can produce better classification results than the traditional autoencoder. This stability benefit seems more significant in the high dimensional, more sophisticated CIFAR-10 dataset compared to the relatively simplistic MNIST dataset. Although the encoder and decoder are shared in the DNA architecture, decoupling between paired bottlenecks creates a meaningful difference between the two reconstruction pathways. As a result, adversarially attacking both pathways is more difficult, even in a white-box scenario. Furthermore, the classification accuracies of the DNA when considering individual pathways are comparable to the conventional autoencoder, and even outperform this baseline on higher magnitude attacks, as seen in Figure 6 and Figure 8.

While the DNA experiments demonstrate the effect of latent feature correlation, they do not demonstrate state-of-the-art performance by themselves. Numerous limiting factors exist, such as the architectures of both the autoencoder and classifiers, which are easier to train, less complex, and do not leverage various data augmentation and regularization techniques when compared to the state-of-the-art. For comparison, the classifier architectures alone achieved 96% on MNIST and 84% on CIFAR-10 datasets, revealing a loss in information during compression. In general, results were less desirable on the CIFAR-10 dataset, which is harder to compress due to less similarity among images and higher dimensionality, which increases the adversarial space. It is also possible that in some attack instances, the autoencoder produced a faithful reconstruction with a small perturbation that instead destabilized the classifier. Generally, however, we observed that adversarial perturbations manifested their effect during the autoencoding process.

The performance gap between the DNA and autoencoder is more pronounced in the CIFAR-10 results, but a variety of scaling challenges should be considered for more complex tasks. Foremost, it is well established that adversarial attacks become dramatically more effective as input dimension increases.[24,25] Complex input domains with high variances over more subspaces are also more difficult to compress in a small latent space. Additionally, implementing the correlations loss requires both QR decomposition and pseudo-inverse computation, both of which are achieved with singular value decomposition (SVD) in most deep learning toolboxes, which can be prohibitively expensive for very large tensors. The batch size is also relevant, as it determines the number of data points used when finding the pseudo-inverse: at least, it must be greater than the bottleneck size, lest the problem is under-constrained. In practice, the batch size should be significantly greater than the bottleneck size to avoid overfitted solutions. In our experiments, a batch size of 500 and 1,250 were used for MNIST and CIFAR-10 experiments, which used bottlenecks of size 64 and 448 respectively.

As stated earlier, changing the architecture of the bottleneck alone did not yield noticeably better results in the MNIST experiments, but increasing the proportion of unshared layers in the model improved performance considerably (Figure 10). This is most likely because perturbations that manifest in shared layers transfer to both pathways; thus, including more shared layers allows transferable attacks to propagate more easily. Architecturally, this suggests a tradeoff between ensemble architectures, where network parameters are entirely separate, and joint architectures with increasingly shared parameters. More unshared layers promise better adversarial robustness, but require more compute during training and inference, and more memory to store. The exact cost and benefit are dependent on the size and depth of shared layers as well as the context of the task.

Additionally, although this method shows how one can decrease transferability of adversarial attacks found using common first-order methods, it does not guarantee adversarial robustness. Many attacks, particularly those with higher magnitude still transfer. The results would suggest, however, that linearly decorrelating the latent features at some point in the models will decrease intersection of adversarial subspaces.

The concept of linear feature correlation between models explored here further explains why adversarial attacks transfer, and how seemingly different models could be extracting the same features. In general, this finding is synergistic with other observations of DNNs, including the existence of multiple equivalent local minima in the parameter space during optimization, and naturally diminished attack transferability in sensible high capacity, highly non-linear models, where deep latent spaces may naturally be non-linearly related, but have weaker linear correlation. [4,6] It would also imply that transferability would be less between models that inherently extract different features due to their architecture: for example CNNs, which focus more on spatially local features, and vision transformers,[26] which have larger receptive fields. While transferability between these two architectures has not been investigated to our knowledge, Paul and Chen show very different adversarial noise patterns between vision transformers and CNNs, with transformer perturbations being more distributed in the frequency domain.[27]

Practically speaking, the concept of correlational loss could be leveraged in a black-box or grey box setting, where one could first traditionally train a surrogate model, and then also train a true, deployable model for the same task but impose a feature correlation loss between the fixed surrogate and the deployment model, presuming that most black box attacks be optimized to a model like the surrogate. It may also be possible to detect adversarial attacks with an ensemble of networks trained with a correlation loss, since it is less likely that these models would reach a consensus on adversarial samples.

Although the current study has not been focused on any specific task, our idea and findings open a new direction to stabilize deep neural networks. Correlation loss can be further explored to determine whether linear correlation or mutual information as a whole is related to transferability. Much more sophisticated architectures can be designed and evaluated. Hopefully, this line of research will promote accurate and stable real-world AI applications.

In this paper, we have demonstrated how adversarial transferability is related to a strong correlation between the features extracted from different models. We have also introduced a novel approach of directly decorrelating these features to reduce attack transferability and demonstrate how even a small pathway deviation along with this decorrelation can reduce transferability of separate reconstructions with a Dual Neck Autoencoder. There are numerous opportunities for further investigations; for example, optimizing this approach and scaling it to larger problems.

## Experimental Procedures

### Resource Availability

#### Lead Contact
Ge Wang, PhD (email: wangg6@rpi.edu)

#### Materials Availability
The deep learning models for the DNA experiments in this work are publicly available via GitHub (https://github.com/WANG-AXIS/DNA).

#### Data and Code Availability
The working code for training and testing models are publicly available via GitHub (https://github.com/WANG-AXIS/DNA).

## Methodology

### Defining Transferability

Consider an ensemble of two classification models $F_i$ and $F_j$, adversarial sample crafted to target model $i$ as $\hat{x}_i \in \hat{X}_i$ with label $y \in Y$. We define the transfer rate as the probability that an adversarial sample will be incorrectly classified by a separate model, given that the sample successfully fools the target model. In other words, $transfer\ rate = P(F_j(\hat{x}_i) \neq y | F_i(\hat{x}_i) \neq y)$

### Training Parametrically Different Models

For testing transferability in parallel MNIST classification models, the architecture in Figure 11 was used. Models $f_1$ and $f_2$ were trained in parallel with batch samples $X$ for 10 epochs using the following objective

$$J(\theta_1, \theta_2, X, Y) = \text{CE}(f_1(\theta_1, X), Y) + \text{CE}(f_2(\theta_2, X), Y) + \lambda(D_t - \|\theta_1 - \theta_2\|_2^2)^2 \quad (5)$$

with CE as the cross-entropy function between logits and labels. To incentivize parametric distance, we added a loss term to encourage a desired Frobenius norm $D_t$ (set to 1000, 1500, 2000, 2500, and 3000 in different trials) between model parameters $\theta_1$ and $\theta_2$ ($\lambda = 10^{-5}$). After training a pair of networks, computed FGSM attacks of varying magnitude $\epsilon$ were computed for one network and tested with the other network for transferability.

### Feature Correlation and Decorrelation Experiments

If $X \in \mathbb{R}^{N \times D}$ is a batch of $N$ input samples, then $Z_1, Z_2 \in \mathbb{R}^{N \times M}$ are vectorized hidden features of length $M$ from models $f_1$ and $f_2$ respectively. The specific hidden layer observed for feature correlation is identified in Figure 11. $Z_1$ and $Z_2$ are relatable by a simple linear regression:

$$Z_2 = \mathbf{Z_1} W \; ; \; \mathbf{Z_1} = [Z_1, 1] \tag{6}$$

where the matrix $W$ contains the optimal regression weights using ordinary least squares. More specifically,

$$R^2 = 1 - \frac{SS_{res}}{SS_{total}} = 1 - \frac{\left\| \left( I - \mathbf{Z_1}(\mathbf{Z_1}^\top \mathbf{Z_1})^{-1} \mathbf{Z_1}^\top \right) Z_2 \right\|_2^2}{\| Z_2 - \bar{Z}_2 \|_2^2} \tag{7}$$

$\bar{Z}_2 \in \mathbb{R}^{N \times M}$ where each row is the sample mean of $Z_2$. Training a decorrelated pair of models is done by simply adding the correlation loss term $\mathcal{L}_R$ to the objective function:

$$J(\theta_1, \theta_2, X, Y) = \text{CE}(f_1(\theta_1, X), Y) + \text{CE}(f_2(\theta_2, X), Y) + \lambda \mathcal{L}_R(f_1, f_2, \theta_1, \theta_2, X) \tag{8}$$

This decorrelating term, is essentially the correlation coefficient of features $Z_1, Z_2$ for a sample batch $X$ at a proper hidden layer in the neural networks. It was found that using the log transform and inserting a small constant value $\varepsilon$ (set to 0.001 in all of our experiments) is beneficial for stability:

$$\mathcal{L}_R = \log(SS_{total} + \varepsilon) - \log(SS_{res} + \varepsilon)$$

$$\mathcal{L}_R = \log(\| Z_2 - \bar{Z}_2 \|_2^2 + \varepsilon) - \log\left( \left\| \left( I - \mathbf{Z_1}(\mathbf{Z_1}^\top \mathbf{Z_1})^{-1} \mathbf{Z_1}^\top \right) Z_2 \right\|_2^2 + \varepsilon \right) \tag{9}$$

Practically speaking, $Z_1$ is not always full column rank. While solutions exist for theoretically more stable rank-deficient pseudo-inverse computations, we found that the most practical method is to first use the QR decomposition to find dependent columns of $Z_1$ and remove them. Algorithm 1 summarizes this process.

```
Algorithm 1. Computation of the correlation loss between latent features.
#N: batch size
#Z_1, Z_2 ∈ ℝ^{N×M}: vectorized batch features from two networks
#ε = 0.001: constant for stability
#η = 0.0005: criteria for determining independent columns

_, R = QR(Z_1) #QR decomposition
column_index = abs(diag(R)/max(diag(R))) > η #find index of independent columns
Z_1 = [Z_1[:, column_index], 1] #remove dependent columns and augment in 1s column
SS_res = ||(I − Z_1(Z_1^⊤ Z_1)^{−1} Z_1^⊤) Z_2||_2^2 #Residual sum of squares from the OLS solution
SS_total = ||Z_2 − Z̄_2||_2^2 #Total sum of squares
ℒ_R = log(SS_total + ε) − log(SS_res + ε)
```

$\lambda$ was set to 0.05 for all experiments involving decorrelation. All network pairs evaluated in Figure *3* were trained using an Adam optimizer for 20 epochs (learning rate: $5e - 3$). $\ell_\infty$ FGSM and PGD attacks were optimized directly using Equations (2) and (3) respectively. $\ell_2$ attacks were optimized in a similar fashion but were modified to be constrained by the $\ell_2$ norm rather than $\ell_\infty$. For the same norm constraint $\epsilon$, $\ell_2$ attacks generate less noise and disruption compared to $\ell_\infty$. Thus, to create comparable adversarial attacks, the norm is scaled by the square root of the input dimension, i.e., the norm constraint is $\epsilon$ for $l_\infty$ attacks but $28\epsilon$ for $l_2$ attacks.

The full theory and implementation of DVERGE can be found in Yang, et al.[22] Our implementation of the approach is similar to that presented in the original paper, but simplified in that 1) Each ensemble only contains two models, 2) distilled features are only computed at a single hidden layer (the same layer used for decorrelation), which is done to reduce variability in the results, 3) momentum is not used when optimizing distilled feature samples.

### DNA Experiments

Part of our goal in testing the DNA is to generate adversarial attacks and observe if the divergent structure and decorrelation mechanism can break transferability between the two reconstructions. Attacks targeting the mean square error of either reconstruction simply results in added background noise in the output rather than differences in semantic content (e.g., causing the autoencoder to reconstruct a perceptually different digit). As such, we attach a classifier $C$ after the DNA and define a classification-based objective:

$$J(F, C, x_i) = \text{CE}(C(F(x_i, \boldsymbol{\theta})_1), y_i)^2 + \text{CE}(C(F(x_i, \boldsymbol{\theta})_2), y_i)^2 \tag{10}$$

This is essentially the squared $L_2$ norm of the cross-entropy losses from the two reconstructions, which is favored over the $L_1$ norm to avoid a sparse solution where only one loss is maximized. As such, this objective attacks both bottlenecks simultaneously, searching for intersections between the two adversarial spaces.

For each dataset, a DNA trained with $\lambda = 0.05$ is compared to a traditional autoencoder of a comparable architecture (i.e., an identically designed encoder, single bottleneck, and decoder in series, with no correlation loss mechanism). Figure 12 and Figure 13 illustrate the exact architectures for the DNA used in MNIST and CIFAR experiments respectively. For each trial, we first train a DNA using the loss in Eq. (4) and then a separate classifier to classify the output images (40 epochs each). After this, we targeted the model using FGSM and PGD (40 iterations) adversarial attacks of magnitude $\epsilon = 0.05, 0.1, 0.15$ for MNIST and $\epsilon = 2/255, 4/255,$ and $8/255$ for CIFAR-10 while recording classification accuracies using the classifier with either DNA output. The two reconstruction pathways were simultaneously attacked using the objective in Eq. (10). The comparison autoencoders were trained in a similar fashion, except a single mean-square error loss was used for training, and the following classification-based objective was used to attack the single reconstruction pathway:

$$J(F, C, x_i) = \text{CE}(C(F(x_i, \boldsymbol{\theta}), y_i) \tag{11}$$

All models were implemented using Pytorch 1.8.1 on an Nvidia Titan RTX GPU with 24GB of VRAM. Adam optimizer was used in all cases.

## Acknowledgments

This work was partially supported by U.S. National Institute of Health (NIH) grants R01EB026646, R01CA233888, R01CA237267, R01HL151561, R21CA264772, and R01EB031102.

## Author Contributions

C.W. and G.W. conceived the idea for this study. C.W. designed the models, executed all experiments, and drafted the paper. G.W. was involved in results analysis, iterative drafting, and editing.

## Declaration of Interests

The authors declare no competing interests.

## References

1. Szegedy, C. *et al.* Intriguing properties of neural networks. *ArXiv13126199 Cs* (2014).
2. Carlini, N. & Wagner, D. Audio Adversarial Examples: Targeted Attacks on Speech-to-Text. *ArXiv180101944 Cs* (2018).
3. Antun, V., Renna, F., Poon, C., Adcock, B. & Hansen, A. C. On instabilities of deep learning in image reconstruction and the potential costs of AI. *Proc. Natl. Acad. Sci.* **117**, 30088–30095 (2020).
4. Papernot, N., McDaniel, P. & Goodfellow, I. Transferability in Machine Learning: from Phenomena to Black-Box Attacks using Adversarial Samples. *ArXiv160507277 Cs* (2016).
5. Goodfellow, I. J., Shlens, J. & Szegedy, C. Explaining and Harnessing Adversarial Examples. *ArXiv14126572 Cs Stat* (2015).

6. Ma, A., Makelov, A., Schmidt, L., Tsipras, D. & Vladu, A. Towards Deep Learning Models Resistant to Adversarial Attacks. 28.
7. Ren, K., Zheng, T., Qin, Z. & Liu, X. Adversarial Attacks and Defenses in Deep Learning. *Engineering* **6**, 346–360 (2020).
8. Kurakin, A., Goodfellow, I. & Bengio, S. Adversarial examples in the physical world. *ArXiv160702533 Cs Stat* (2017).
9. Carlini, N. & Wagner, D. Defensive Distillation is Not Robust to Adversarial Examples. *ArXiv160704311 Cs* (2016).
10. Papernot, N., McDaniel, P., Wu, X., Jha, S. & Swami, A. Distillation as a Defense to Adversarial Perturbations against Deep Neural Networks. *ArXiv151104508 Cs Stat* (2016).
11. Carlini, N. & Wagner, D. Towards Evaluating the Robustness of Neural Networks. *ArXiv160804644 Cs* (2017).
12. Papernot, N. *et al.* The Limitations of Deep Learning in Adversarial Settings. *ArXiv151107528 Cs Stat* (2015).
13. Song, Y., Shu, R., Kushman, N. & Ermon, S. Constructing Unrestricted Adversarial Examples with Generative Models. *ArXiv180507894 Cs Stat* (2018).
14. Lee, H., Han, S. & Lee, J. Generative Adversarial Trainer: Defense to Adversarial Perturbations with GAN. *ArXiv170503387 Cs Stat* (2017).
15. Athalye, A., Carlini, N. & Wagner, D. Obfuscated Gradients Give a False Sense of Security: Circumventing Defenses to Adversarial Examples. *ArXiv180200420 Cs* (2018).
16. Liu, X. & Hsieh, C.-J. Rob-GAN: Generator, Discriminator, and Adversarial Attacker. *ArXiv180710454 Cs Stat* (2019).
17. Chaubey, A., Agrawal, N., Barnwal, K., Guliani, K. K. & Mehta, P. Universal Adversarial Perturbations: A Survey. *ArXiv200508087 Cs* (2020).
18. Tramèr, F., Papernot, N., Goodfellow, I., Boneh, D. & McDaniel, P. The Space of Transferable Adversarial Examples. *ArXiv170403453 Cs Stat* (2017).
19. Moosavi-Dezfooli, S.-M., Fawzi, A., Fawzi, O. & Frossard, P. Universal adversarial perturbations. *ArXiv161008401 Cs Stat* (2017).
20. Hirano, H., Minagi, A. & Takemoto, K. Universal adversarial attacks on deep neural networks for medical image classification. *BMC Med. Imaging* **21**, 9 (2021).
21. Ilyas, A. *et al.* Adversarial Examples Are Not Bugs, They Are Features. *ArXiv190502175 Cs Stat* (2019).
22. Yang, H. *et al.* DVERGE: Diversifying Vulnerabilities for Enhanced Robust Generation of Ensembles. *ArXiv200914720 Cs Stat* (2020).
23. Mikolov, T., Chen, K., Corrado, G. & Dean, J. Efficient Estimation of Word Representations in Vector Space. *ArXiv13013781 Cs* (2013).
24. Dube, S. High Dimensional Spaces, Deep Learning and Adversarial Examples. *ArXiv180100634 Cs* (2018).
25. Gilmer, J. *et al.* Adversarial Spheres. *ArXiv180102774 Cs* (2018).
26. Dosovitskiy, A. *et al.* An Image is Worth 16x16 Words: Transformers for Image Recognition at Scale. *ArXiv201011929 Cs* (2021).
27. Paul, S. & Chen, P.-Y. Vision Transformers are Robust Learners. *ArXiv210507581 Cs* (2021).

# Figure Legends

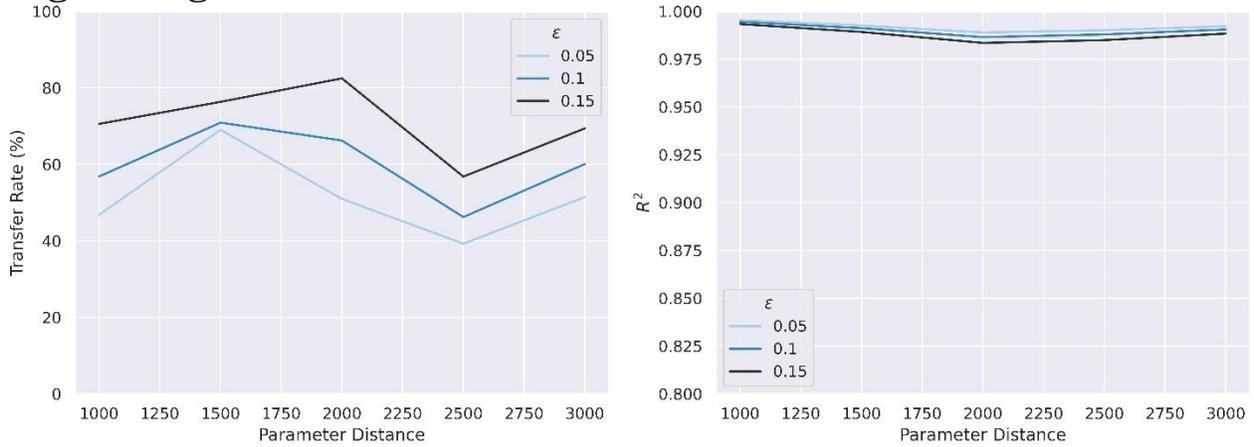

Figure 1: Semantically same encoding with different distances in the space of network parameters. (Left): Attack transferability rate vs parametric distance between network pairs. Transferability is measured as the proportion of adversarial attacks from one model that successfully cause misclassification on the opposing network. Parameter distance is measured as the square Frobenius norm $\|\theta_1 - \theta_2\|_2^2$. (Right): Correlation coefficient between vectorized latent features extracted from either model while processing adversarial samples. $\epsilon$ is the FGSM attack magnitude.

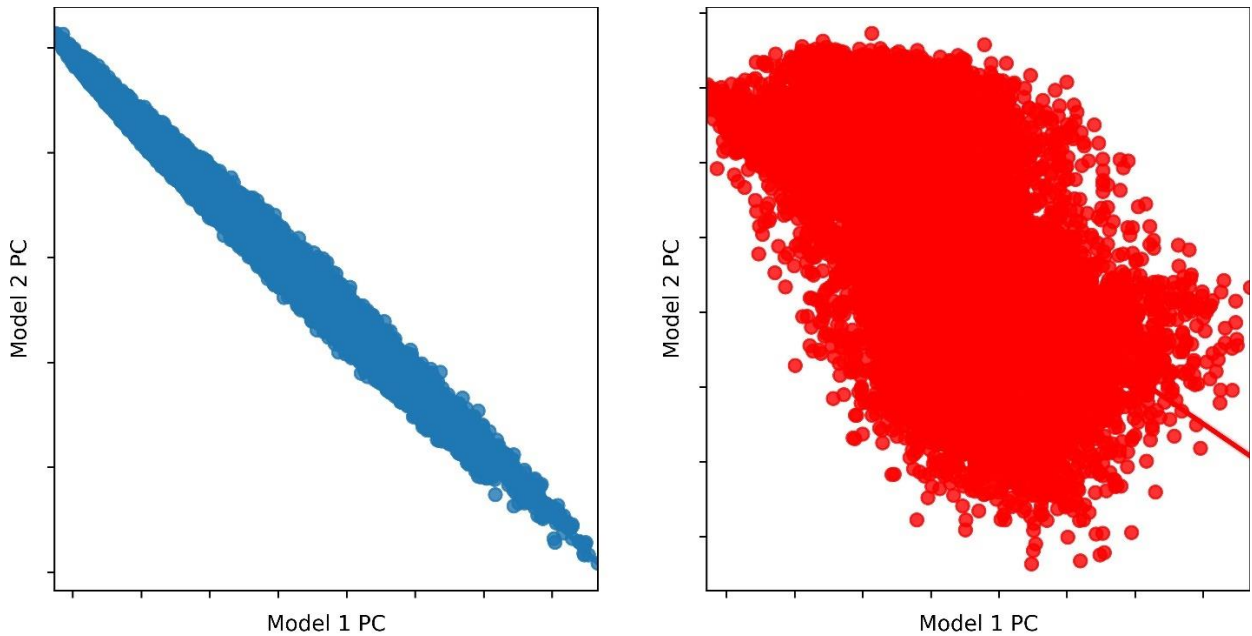

Figure 2: Comparison of hidden feature distributions. Left: Scatter plot of features extracted from samples via two models trained in parallel ($R^2 = 0.991$). Right: Scatter of sample features extracted from two models trained with decorrelation loss ($R^2 = 0.605$). Extracted features for each model were compressed into one dimension by projecting along the largest principal component.

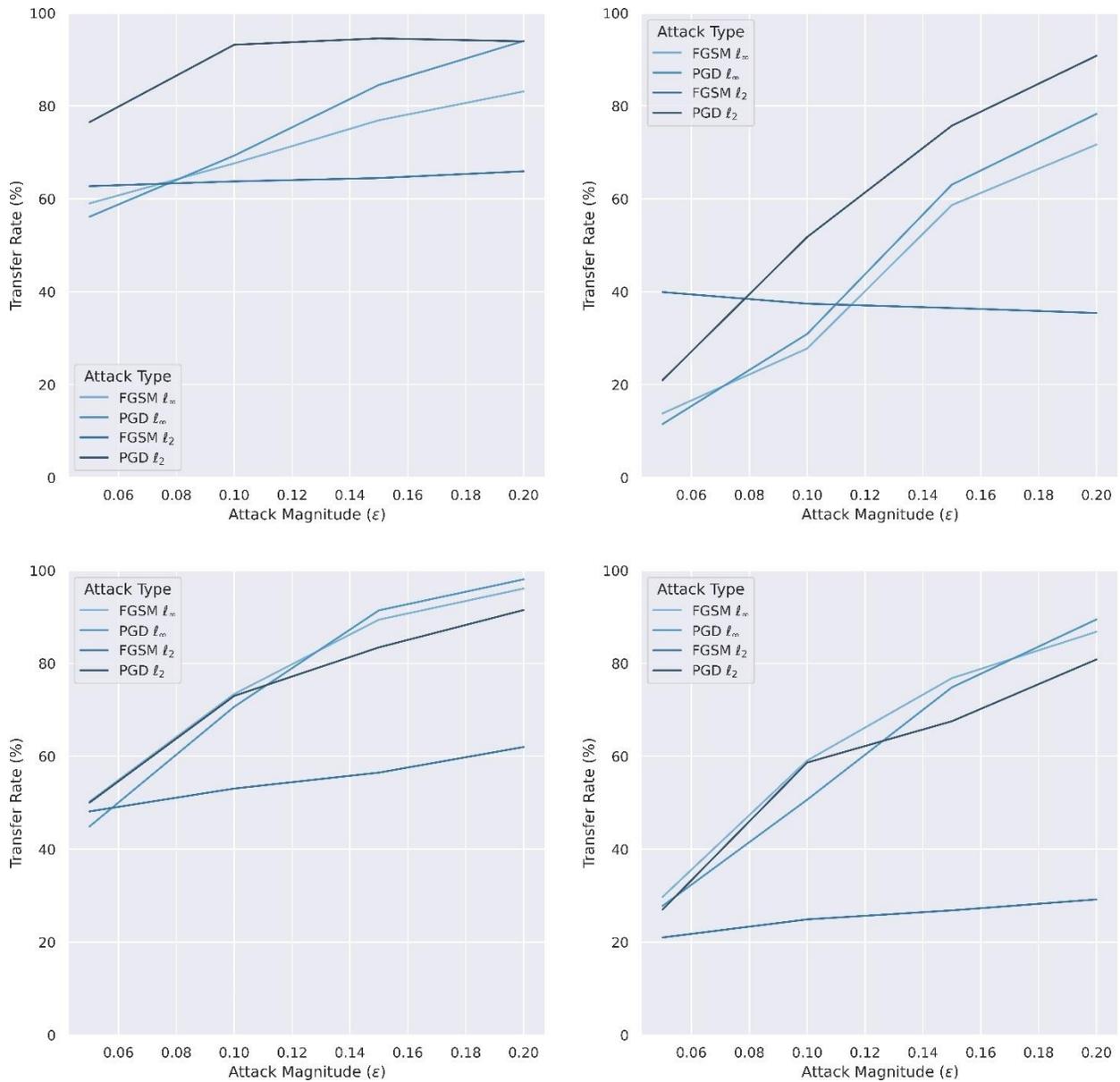

Figure 3: Adversarial transference using regular parallel models vs decorrelated models. Top left: transfer rate between parallel CNNs (average natural accuracy: 94.5%). Top right: transfer rate between decorrelated CNNs (average natural accuracy: 95.4%). Bottom left: transfer rates between parallel fully connected (FC) models (average natural accuracy: 98.2%). Bottom right: transfer rates between decorrelated FC models (average natural accuracy: 96.2%).

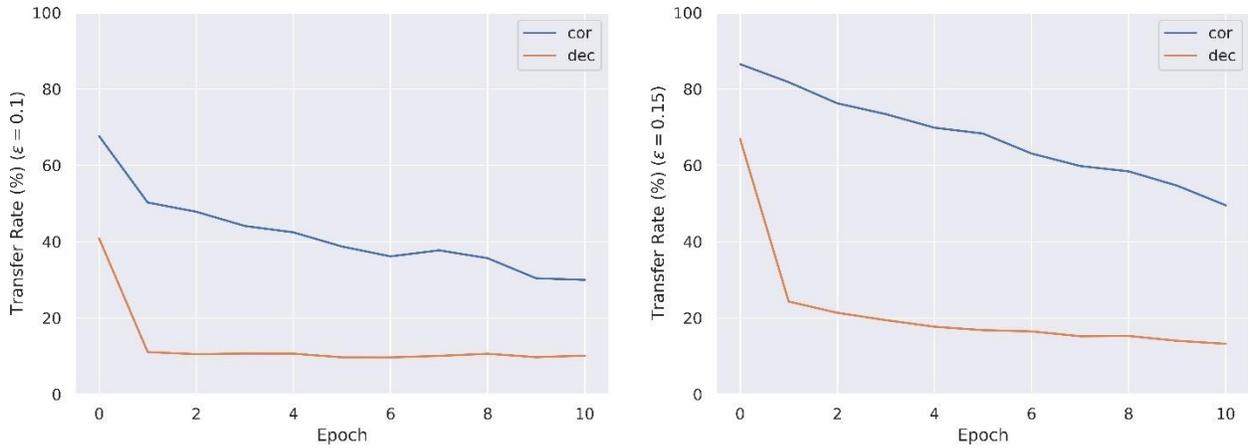

Figure 4: Attack transfer rates with DVERGE: Transfer rates of attacks between model pairs diversified using 10 epochs of DVERGE training. cor: model pair initialized with regular training. dec: model pair initialized with decorrelation loss.

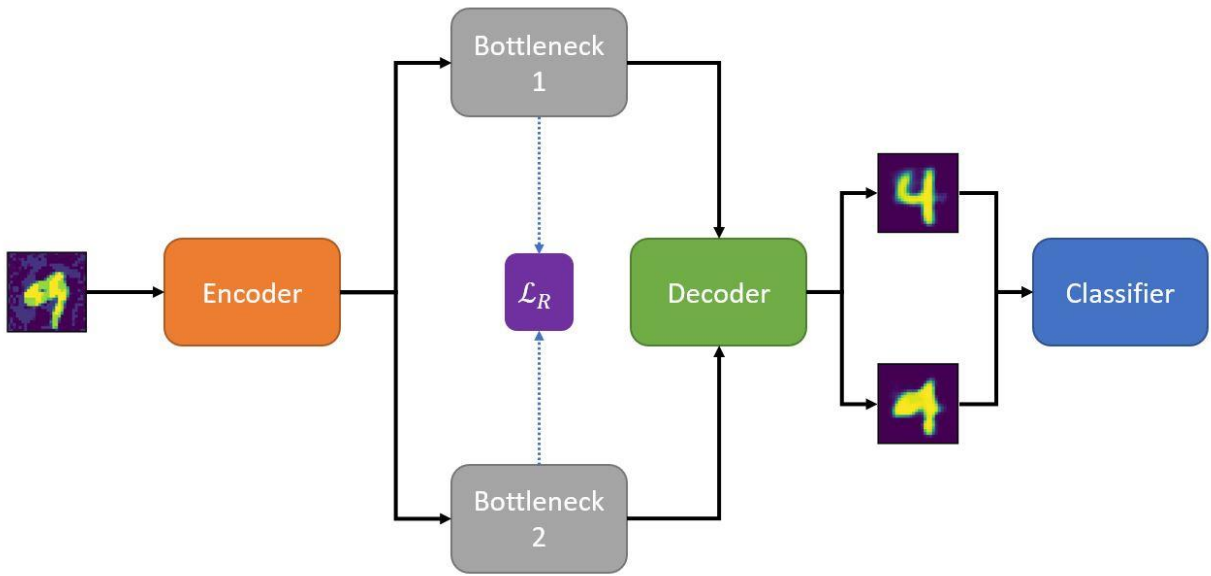

Figure 5: Proposed dual neck autoencoder (DNA). A common encoder and decoder are shared to compress and decompress information respectively. Two separate bottleneck modules differently encode information into fully compressed states which are decorrelated using the correlation loss term. The decoder produces a separate reconstruction from each encoding.

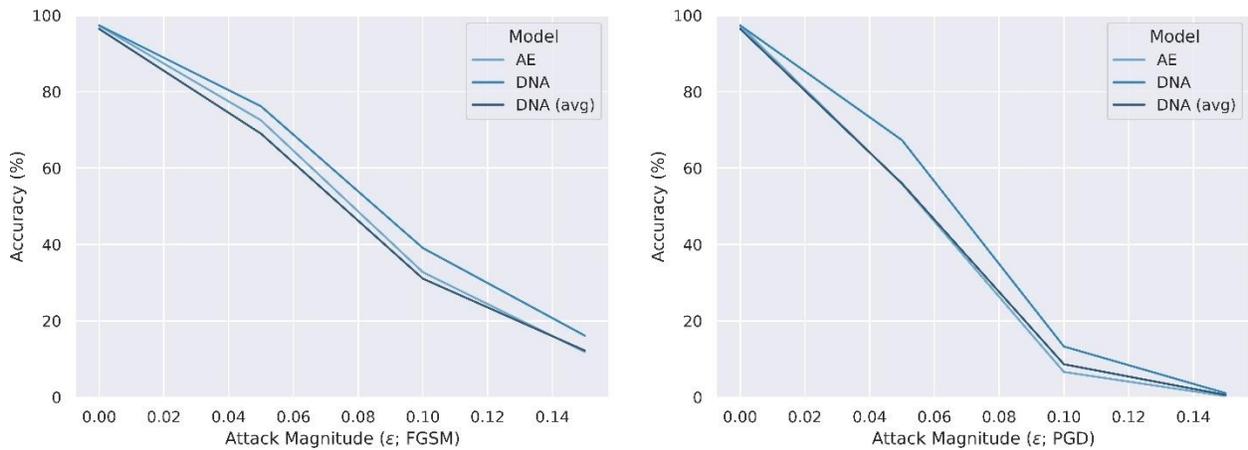

Figure 6: Reconstruction accuracies of the DNA and traditional autoencoder (AE) coupled with the CNN classifier on MNIST dataset with FGSM (left) and PGD (right) attacks of varying magnitude $\epsilon$. DNA (avg) is the average accuracy between individual reconstruction paths in the DNA.

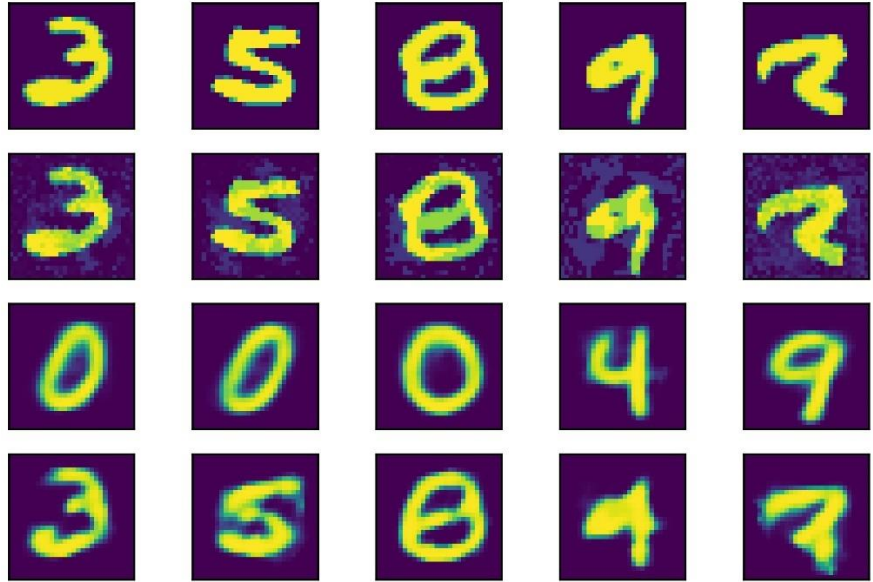

Figure 7: Example reconstructions of classification-based strong adversarial (PGD, $\epsilon = 0.15$) attacks from DNA. 1[st] row: Natural samples. 2[nd] row: Perturbed samples. 3[rd] row: Incorrect reconstruction from one bottleneck. 4[th] row: Correct reconstruction from alternative bottleneck.

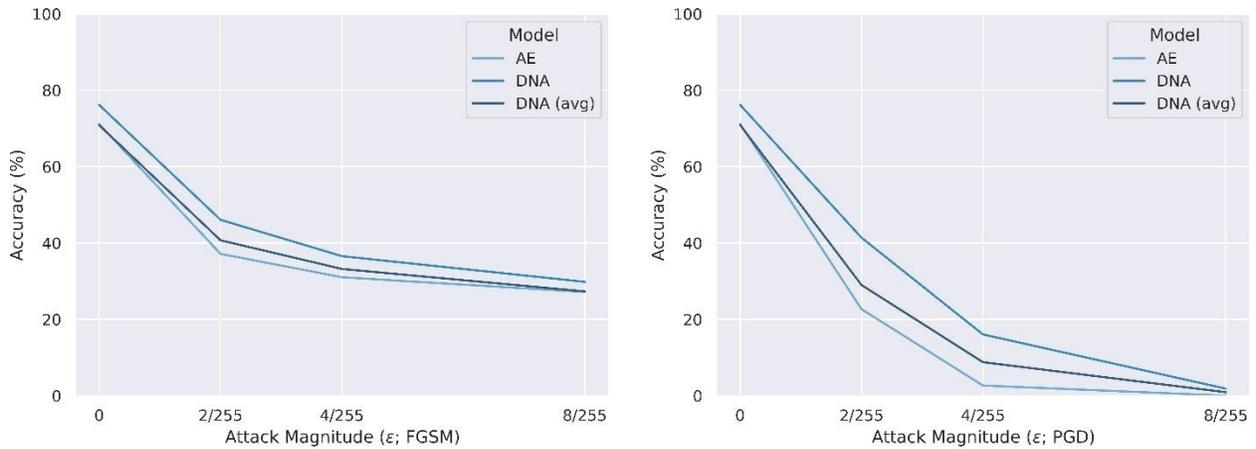

Figure 8: Reconstruction accuracies of the DNA and traditional autoencoder (AE) coupled with the CNN classifier on CIFAR-10 dataset with FGSM (left) and PGD (right) attacks of varying magnitude $\epsilon$. DNA (avg) is the average accuracy between individual reconstruction paths in the DNA.

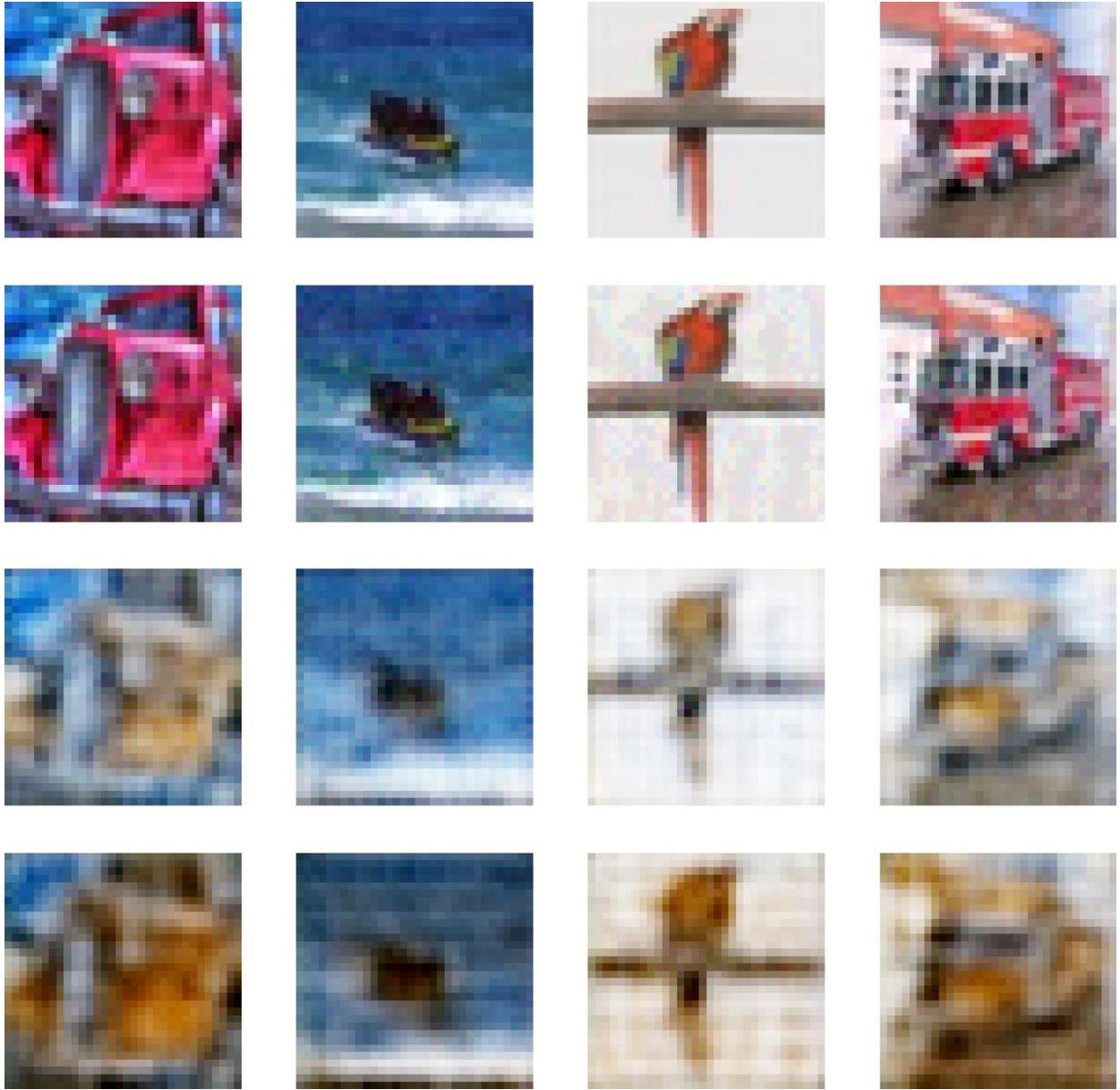

Figure 9: Example reconstructions of classification-based strong adversarial (PGD, $\epsilon = 8/255$) attacks from DNA on CIFAR-10 data. 1st row: Natural samples. 2nd row: Perturbed samples. 3rd row: Incorrect reconstruction from one bottleneck. 4th row: Correct reconstruction from alternative bottleneck.

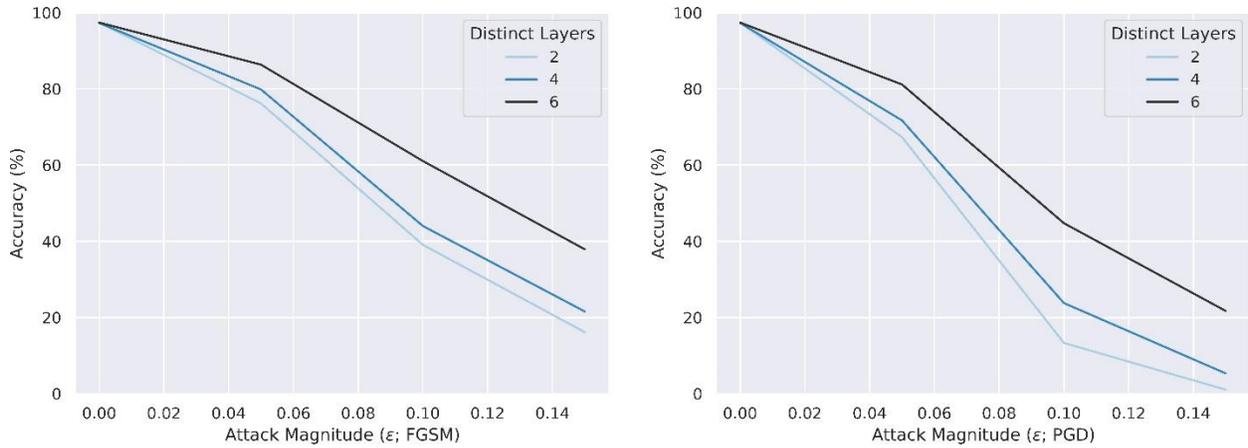

Figure 10: Reconstruction accuracies of the DNA with varying proportions of unshared layers on the MNIST dataset. Legend indicates the number of network layers unshared between the two reconstruction pathways. FGSM (left) and PGD (right) attacks of varying magnitude $\epsilon$ were used.

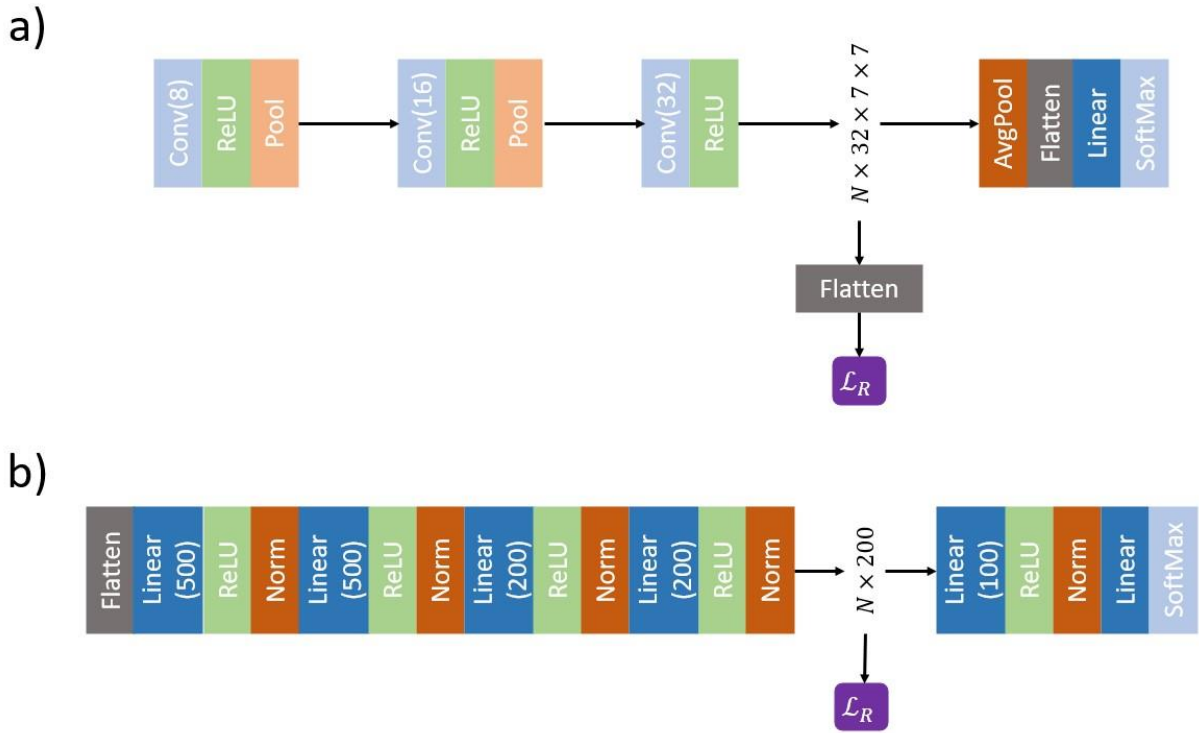

Figure 11: Architectures of the classifier models used to test correlation loss $\mathcal{L}_R$. a) CNN, b) Fully Connected (FC). Conv(K): 3x3, 1-stride, zero-padded 2D convolutional layer with K output filters. Pool: 2x2 max pooling. AvgPool: 7x7 average feature pooling. Linear: Densely connected layer with size 10 output. Note that two models of the same architecture were trained in parallel, simultaneously inputting flattened features for comparison and $\mathcal{L}_R$ calculation.

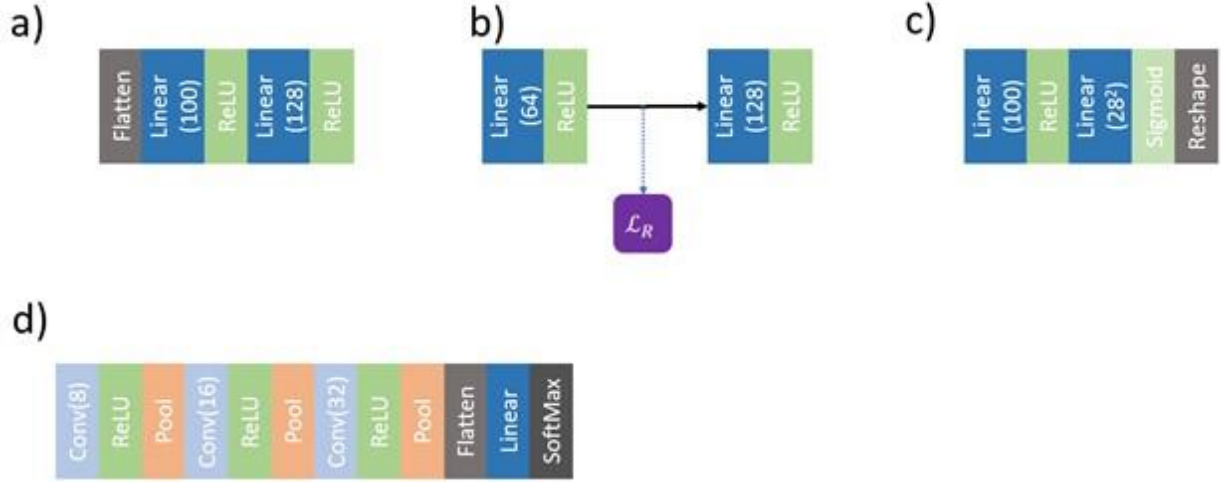

Figure 12: Architectures used in the DNA experiments with the MNIST digit dataset. a) encoder, b) bottleneck, c) decoder d) classifier. Conv(K): 3x3, 1-stride, zero-padded 2D convolutional layer with K output filters. Pool: 2x2 max pooling. Linear(K): Densely connected layer with size K (10 if not specified) output.

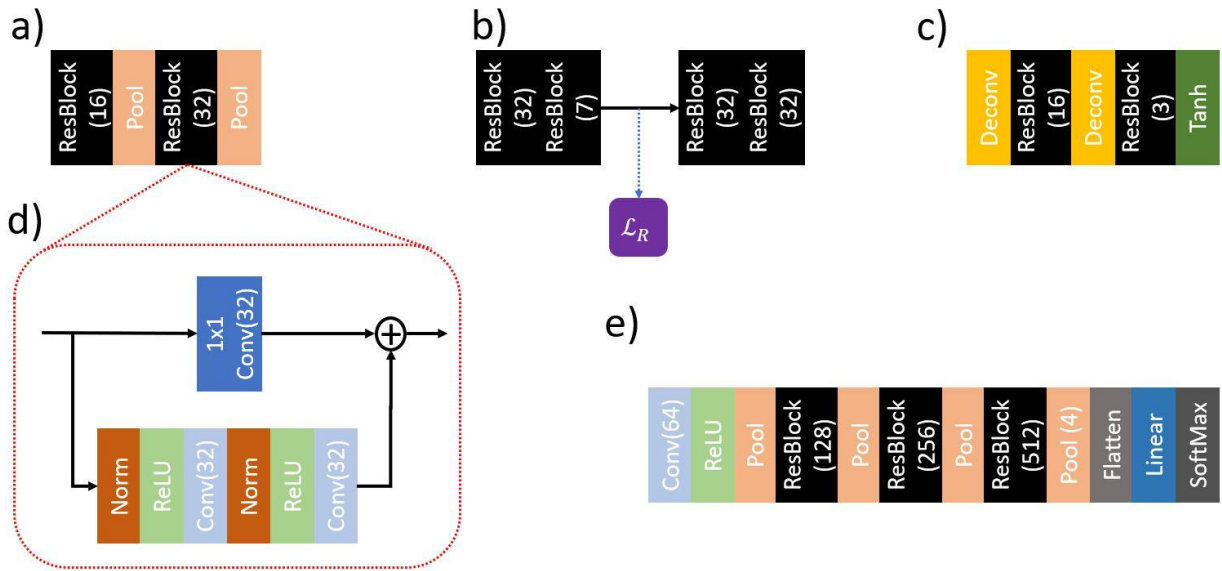

Figure 13: Architecture used in the DNA experiments with the CIFAR-10 dataset. a) encoder, b) bottleneck, c) decoder, and d) residual block, e) classifier. Conv(K): 3x3, 1-stride, zero-padded 2D convolutional layer with K output filters. Pool: 2x2 max pooling. Norm: Batch normalization layer. Deconv: 2x2 transpose convolution layer (number of input and output channels are equivalent). 1x1 Conv(K): 1x1 2D convolution with K output filters. Linear: Densely connected layer with size 10 output.